\documentclass[10pt,twocolumn,letterpaper]{article}

\usepackage{wacv}              % To produce the CAMERA-READY version
\definecolor{wacvblue}{rgb}{0.21,0.49,0.74}
\usepackage[pagebackref,breaklinks,colorlinks,allcolors=wacvblue]{hyperref}
\usepackage[ruled,vlined]{algorithm2e}
\usepackage{multirow}
\usepackage{adjustbox}

\def\wacvPaperID{671} % *** Enter the WACV Paper ID here
\def\confName{WACV}
\def\confYear{2026}

\title{MMCM: Multimodality-aware Metric using Clustering-based Modes \\ for Probabilistic Human Motion Prediction}

\author{Kyotaro Tokoro, Hiromu Taketsugu, Norimichi Ukita\\
Toyota Technological Institute\\
{\tt\small \{sd24439,sd25502,ukita\}@toyota-ti.ac.jp}
}

\begin{document}
\maketitle
\begin{abstract}
This paper proposes a novel metric for Human Motion Prediction (HMP). 
Since a single past sequence can lead to multiple possible futures,
a probabilistic HMP method predicts such multiple motions.
While a single motion predicted by a deterministic method is evaluated only with the difference from its ground truth motion, multiple predicted motions should also be evaluated based on their distribution. For this evaluation, this paper focuses on the following two criteria. \textbf{(a) Coverage}: motions should be distributed among multiple motion modes to cover diverse possibilities. \textbf{(b) Validity}: motions should be kinematically valid as future motions observable from a given past motion.
However, existing metrics simply appreciate widely distributed motions even if these motions are observed in a single mode and kinematically invalid.
To resolve these disadvantages, this paper proposes a Multimodality-aware Metric using Clustering-based Modes (MMCM).
For (a) coverage, MMCM divides a motion space
into several clusters, each of which is regarded as a mode. These modes are used to explicitly evaluate whether predicted motions are distributed among multiple modes.
For (b) validity, MMCM identifies valid modes by collecting possible future motions from a motion dataset.
Our experiments validate that our clustering yields sensible mode definitions and that MMCM accurately scores multimodal predictions.
Code: \url{https://github.com/placerkyo/MMCM}
\end{abstract}

\section{Introduction}
\label{sec:intro}

\begin{figure}
  \begin{center}
     \includegraphics[width=\linewidth]{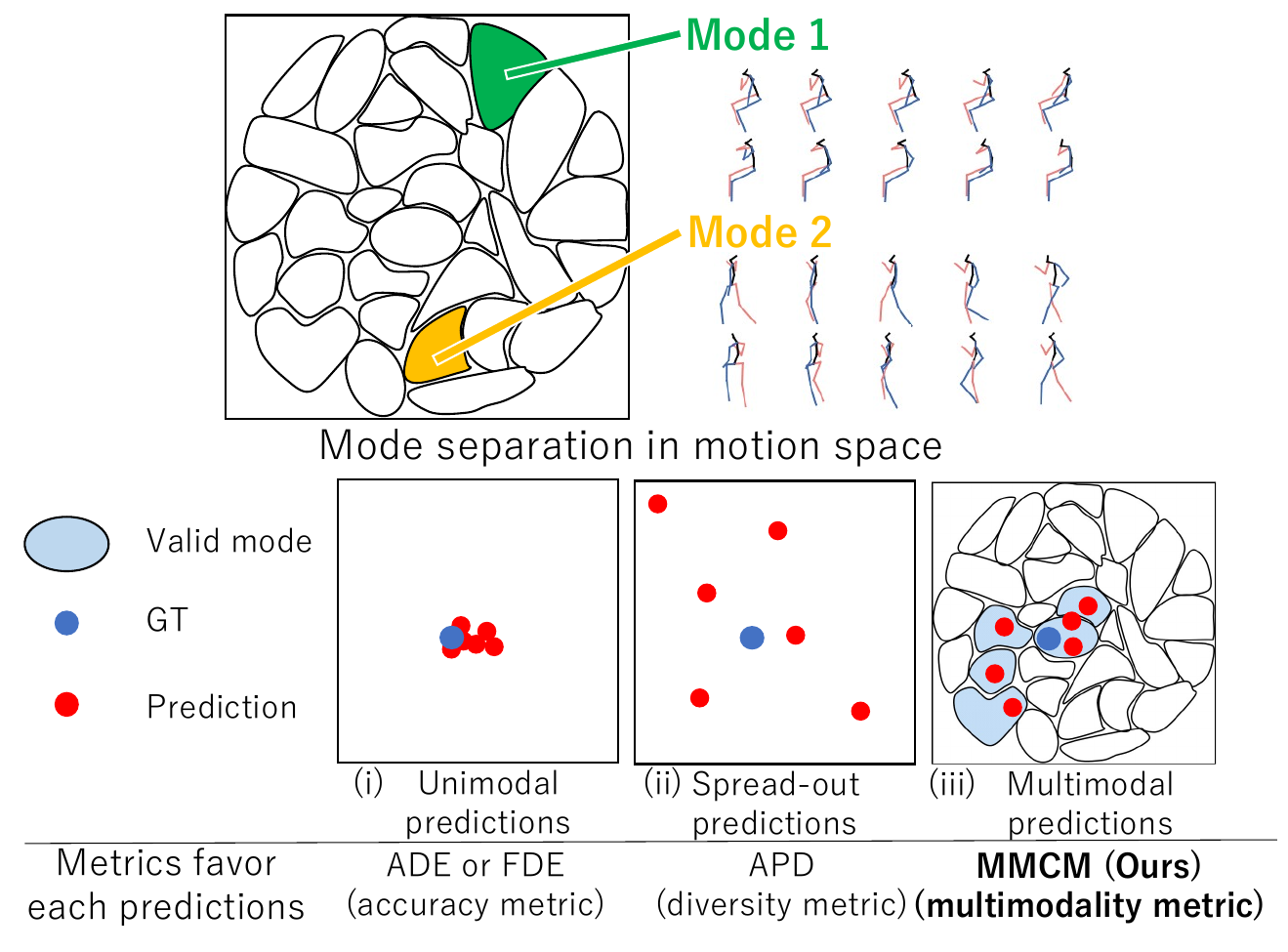}
     \caption[Why a new multimodality metric is needed.]{Why a new multimodality metric is needed. Three prediction patterns are sketched in latent ``motion mode space'' (a motion mode is a set of similar actions): (i) Unimodal (all samples in a few valid modes. Valid mode is a mode that is valid as a continuations of the past), (ii) Spread-out (samples dispersed without regard to the valid modes), and (iii) Multimodal (samples distributed over several valid modes). The table at the bottom visualizes which prediction pattern each metric scores high.}
     \label{fig:ade_apd_ours}
   \end{center}
   \vspace*{-2mm}
\end{figure}

Human Motion Prediction (HMP) provides future motions from a past motion. HMP is applicable to various tasks, \eg, autonomous driving~\cite{DBLP:journals/tiv/PadenCYYF16}, human-robot interaction~\cite{DBLP:journals/ral/TeramaeNM18, DBLP:conf/icra/ChenLKA19, DBLP:conf/iros/GuiZWLMV18, DBLP:conf/iros/KoppulaS13}, and animation~\cite{DBLP:journals/cgf/WelbergenBERO10}.

Compared with similar tasks such as human trajectory prediction~\cite{Trajectronpp,DBLP:conf/iccv/0001U23,SceneTransformer,EmLoco,socialtransmotion}, HMP is challenging because it is inherently ill-posed: given a past motion, there may be many valid future motions, as predicted by probabilistic HMP methods~\cite{DBLP:conf/eccv/YuanK20,DBLP:conf/iccv/BarqueroEP23,DBLP:conf/eccv/SunC24,DBLP:conf/iccv/ChenZLPXL23,DBLP:conf/icra/SaadatnejadRMMRMA23,DBLP:conf/aaai/WeiSLLLSH23,DBLP:journals/corr/abs-2501-06035,DBLP:journals/corr/abs-2505-13140}.
However, even if multiple motions are predicted, only one future motion is available as the Ground Truth (GT) future motion of the past motion.
Among these multiple predictions, only the one closest to this GT is evaluated in standard HMP metrics, such as Average Displacement Error (ADE) and Final Displacement Error (FDE).
Since this single GT is insufficient to evaluate whether the predicted motions cover the possible motions, HMP also needs other evaluation metrics.

As such a metric in the literature~\cite{DBLP:conf/eccv/YuanK20,DBLP:conf/iccv/BarqueroEP23,DBLP:conf/eccv/SunC24,DBLP:conf/iccv/ChenZLPXL23,DBLP:conf/icra/SaadatnejadRMMRMA23,DBLP:conf/aaai/WeiSLLLSH23,DBLP:journals/corr/abs-2501-06035}, Average Pairwise Distance (APD)~\cite{DBLP:conf/eccv/YuanK20} evaluates the variance of predicted motions because their diversity should be larger to cover possible future motions.
However, APD cannot assess multimodality of the predicted motions. Since diverse future motions can be not only unimodal (\ie, all future motions are similar and vary gradually) but also multimodal (\ie, several future motions change significantly, such as ``stopping from walking'' rather than ``keep walking''), an HMP metric should evaluate multimodality.
Furthermore, APD can be erroneously large even if kinematically invalid motions are predicted.
On the other hand, this paper focuses on the following two requirements for evaluating multimodality in predicted motions:
(a) \textbf{coverage}, the ability to explicitly assess whether the predicted motions are distributed to multiple modes, and
(b) \textbf{validity}, the ability to evaluate whether the predicted motions fall only within modes that are observable from a given past motion. Such modes are called valid modes in this paper.

As a metric satisfying (a) and (b), this paper proposes Multimodality-aware Metric using Clustering-based Modes (\textbf{MMCM}).
MMCM differs from standard HMP metrics, such as accuracy and diversity metrics (Fig.~\ref{fig:ade_apd_ours}), as follows.
Accuracy metrics, which compare each prediction only to a single GT, favor unimodal predictions in which the predictions collapse onto a few modes (Fig.~\ref{fig:ade_apd_ours} (i)). Diversity metrics such as APD, conversely, reward excessively spread-out predictions that may include abnormal motions that are not valid as future motions of a given past motion (Fig.~\ref{fig:ade_apd_ours} (ii)).
In contrast, the proposed MMCM evaluates multimodality of predictions that cover all valid modes (Fig.~\ref{fig:ade_apd_ours} (iii)).

To quantify multimodality, multiple modes in the motion space are useful cues.
In MMCM, these modes are obtained by dividing the motion space into multiple clusters, each of which is regarded as a mode.
Given these modes, MMCM identifies valid modes for the past motion.
To identify valid modes, MMCM employs MultiModal GTs (MMGTs)~\cite{DBLP:conf/eccv/YuanK20,DBLP:conf/corr/abs-2412-18883}, which are multiple pseudo future motions for a single past. We assume that MMGTs fully cover the range of possible futures if the dataset is sufficiently large.
With the valid modes, MMCM measures how \textbf{(a) coverage} and \textbf{(b) validity} are satisfied in a balanced way.

Our contributions are summarized as follows:
\begin{enumerate}
\item We define motion modes through clustering and evaluate whether predicted motions cover all valid modes, addressing the requirement~\textbf{(a)}.
\item By leveraging our mode definition together with MMGTs, we assess the validity of each prediction, addressing the requirement~\textbf{(b)}.
\item Comprehensive experiments on two large datasets with multiple HMP methods demonstrate that our MMCM offers a sound mode definition and provides a reliable evaluation of multimodality.
\end{enumerate}

\section{Related Work}
\label{section:related_work}

\subsection{Human Motion Prediction     (HMP)}
\label{section:related_methods}

HMP employs a variety of temporal modeling approaches~\cite{DBLP:journals/ijon/LyuCLZW22,DBLP:journals/ivc/DengS24}, including RNNs~\cite{DBLP:conf/iccv/FragkiadakiLFM15, DBLP:conf/cvpr/MartinezB017}, Transformers~\cite{DBLP:conf/3dim/AksanKCH21, DBLP:conf/iccvw/Martinez-Gonzalez21} and GNNs~\cite{DBLP:conf/aaai/YanXL18, DBLP:conf/cvpr/ZhongHZYX22}. However, they are deterministic, yielding only a single motion for a past motion.

A principal way to obtain multiple predictions is to use probabilistic models. 
A multimodality of probabilistic models have been explored, including VAE~\cite{DBLP:conf/eccv/YuanK20,DBLP:conf/mm/DangNLZL22,DBLP:conf/iccv/MaoLS21,DBLP:conf/eccv/XuWG22}, GAN~\cite{DBLP:conf/cvpr/BarsoumKL18,DBLP:conf/cvpr/GurumurthySB17}, probabilistic latent variable~\cite{DBLP:conf/cvpr/SalzmannPR22}, and, diffusion models~\cite{DBLP:conf/iccv/BarqueroEP23,DBLP:conf/eccv/SunC24,DBLP:conf/iccv/ChenZLPXL23,DBLP:conf/icra/SaadatnejadRMMRMA23,DBLP:conf/aaai/WeiSLLLSH23,DBLP:journals/corr/abs-2501-06035}.
For example, DLow~\cite{DBLP:conf/eccv/YuanK20} samples latent codes from multiple Gaussian priors.
In~\cite{DBLP:conf/corr/abs-2412-18883}, multimodal predictions are initially estimated in a low-dimensional feature space, and each estimated feature is employed for HMP in the original motion space.
To evaluate these HMP methods, multimodality metrics are vital.

\subsection{Metrics for Human Motion Prediction}
\label{section:related_metrics}

As with metrics for human pose estimation~\cite{DBLP:conf/cvpr/AndrilukaPGS14,DBLP:conf/eccv/LinMBHPRDZ14,DBLP:conf/mva/MorikiTU25}, several HMP metrics are proposed.
The metrics employed in HMP can be grouped into three categories~\cite{zhu2023human}: accuracy metrics, realism metrics that assess motion validity, and diversity metrics that measure variation across predictions.

The most common accuracy metrics are ADE and FDE. ADE computes the mean $\ell_2$ distance between each predicted joint position and the corresponding GT joint over all frames, whereas FDE measures the $\ell_2$ distance at the final frame only.
To multiple predictions, accuracy is often measured with MultiModal ADE (MMADE) and MultiModal FDE (MMFDE)~\cite{DBLP:conf/eccv/YuanK20}, which respectively take ADE and FDE between each MMGT and its closest prediction.

For assessing motion realism, two widely used metrics are Fr\'{e}chet Inception Distance (FID)~\cite{DBLP:conf/mm/GuoZWZSDG020} and Cumulative Motion Distribution (CMD)~\cite{DBLP:conf/iccv/BarqueroEP23}.
FID uses a pre-trained action classifier to embed entire sequences to measure the Fr\'{e}chet distance between the feature distributions of GT sequences and the predicted ones.
CMD evaluates whether the velocity profile is physically valid by computing the displacement between every pair of consecutive frames.

As a diversity metric, APD~\cite{DBLP:conf/eccv/YuanK20} measures the variation across multiple predictions. APD measures the mean $\ell_2$ distance over all pairs of predictions.
A derivative metric of APD, APDE~\cite{DBLP:conf/iccv/BarqueroEP23}, is obtained by taking the difference between the APD of the predictions and the APD of MMGTs.

None of the existing metrics mentioned above directly measures multimodality; no mode definition is used in these metrics.
By contrast, our MMCM explicitly defines modes, satisfying \textbf{(a) coverage}, and then evaluates whether each prediction falls into one of the valid modes, thereby penalizing abnormal motions and satisfying ~\textbf{(b) validity}.

\section{Problem Statement}
\label{section:problem_statement}

An HMP method receives a past motion sequence of $B$ observed frames to predict the subsequent $T$ future frames. 
Since a single past sequence can lead to multiple futures, recent HMP methods predict $I (> 1)$ candidate future motions for each past sequence.

Throughout this paper, we denote an observed past motion as $\mathbf{X} = \{p_{t-B}, \dots, p_{t-2}, p_{t-1}\}$
and a future motion as $\mathbf{Y} = \{p_t, p_{t+1}, \dots, p_{t+T-1}\}$.
$p_{t}$ represents a human pose at time $t$, expressed by a set of keypoint coordinates.
$I$ candidate futures are written as $\hat{\mathbf{Y}}^i = \{\hat{p}_t^i, \hat{p}_{t+1}^i, \dots, \hat{p}_{t+T-1}^i\}$, where $i \in \{1, \dots, I\}$.

Following prior work~\cite{DBLP:conf/eccv/YuanK20,DBLP:conf/iccv/BarqueroEP23}, given a motion dataset, our method constructs MMGTs by grouping sequences whose past poses are similar to each other. 
Concretely, given a test motion consisting of its past and future sequences, the $\ell_2$ distance between the past sequences of the given motion and each motion in a motion dataset is computed.
If the distance falls below a threshold, the future sequence of that motion is regarded as an MMGT of the past sequence of the given motion.
By collecting MMGTs from all motions in the dataset, a set of MMGTs for the given past sequence is obtained.

\begin{figure*}[t]
  \begin{center}
     \includegraphics[width=\linewidth]{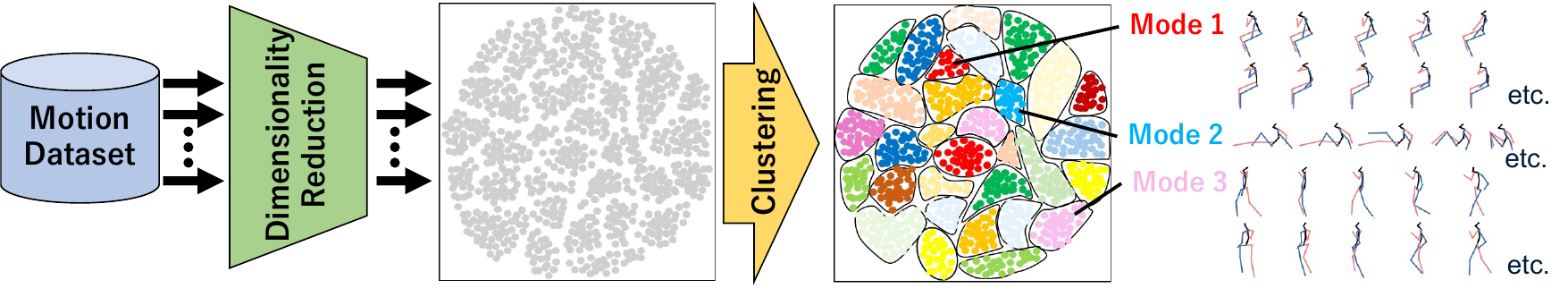}
     \vspace*{-6mm}
     \caption[Clustering-based mode definition]{A large motion dataset is first passed through the Dimensionality Reduction. On this n-dimensional plane, clustering detects high-density regions. Each resulting cluster is recorded as a motion mode (Mode 1, Mode 2, Mode 3, $\cdots$).}
     \label{fig:clustering-based_mode_definition}
   \end{center}
   \vspace*{-3mm}
\end{figure*}

\begin{figure*}[t]
  \begin{center}
     \includegraphics[width=\linewidth]{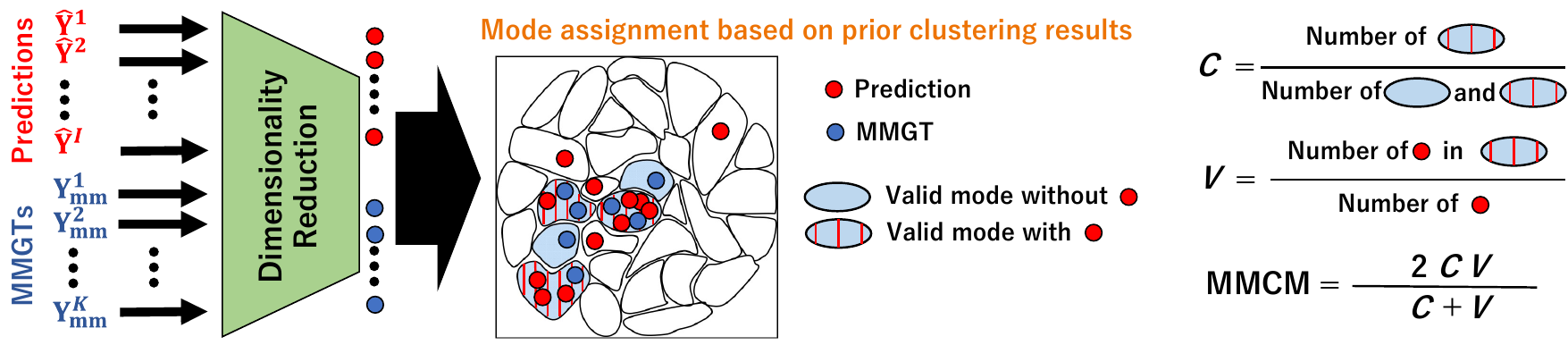}
     \caption[MMCM computation]{Evaluation by MMCM. Both the predictions (top) and the MMGTs (bottom, denoted as $\mathbf{Y}^{k}_{\mathrm{mm}}$) are independently passed through a dimensionality reduction to obtain their latent embedding. Each embedded sequence is assigned to its nearest mode obtained from the motion space clustering. 
     Among all the modes, those including MMGT embeddings are regarded as valid modes. MMCM is defined as the harmonic mean of $C$ and $V$, as shown on the right side of the figure.
     }
     \label{fig:evaluation_metric_computation}
   \end{center}
\end{figure*}

\section{Method}
\label{section:method}

MMCM assesses predictions in two stages. Stage~1 is clustering-based mode definition (Sec.~\ref{subsection:clustering-based_mode_definition}), as illustrated in Fig.~\ref{fig:clustering-based_mode_definition}.  
Stage~2 is Multimodality metric evaluation (Sec.~\ref{subsection:MMCM_computation_for_predictions}), as illustrated in Fig.~\ref{fig:evaluation_metric_computation}.

\subsection{Stage~1: Clustering-based Mode Definition}
\label{subsection:clustering-based_mode_definition}

Quantifying multimodality requires two steps: (1) defining motion modes and (2) determining which mode each prediction belongs to. 
For MMCM, a motion space constructed from a large training dataset is partitioned into clusters of similar motions.
Each cluster is regarded as a mode (e.g., Mode 1, Mode 2, and Mode 3 in Fig.~\ref{fig:clustering-based_mode_definition}).
Our clustering-based mode definition comprises two modules: ``Dimensionality Reduction of Motion'' (green part in Fig.~\ref{fig:clustering-based_mode_definition}) and ``Clustering in Motion Space'' (yellow part in Fig.~\ref{fig:clustering-based_mode_definition}).

\vspace*{-5mm}
\paragraph{Dimensionality Reduction of Motions.}

In the original high-dimensional motion space (\ie, \(\text{3D coordinate axes} \times \text{keypoints} \times \text{frames}\)), motion data may be sparsely distributed. 
In such a high-dimensional space, it is hard to detect cluster boundaries. 
This problem is alleviated by representing the motion data in its low-dimensional latent space.
For dimensionality reduction, we adopt a two-stage pipeline in which an autoencoder is followed by UMAP~\cite{DBLP:journals/jossw/McInnesHSG18}, because existing methods~\cite{DBLP:journals/corr/abs-2501-07729, DBLP:conf/icpr/McConvilleSPC20, DBLP:journals/neco/SainburgMG21} demonstrate its effectiveness as a preprocess for clustering.

The autoencoder provides a compact latent representation that retains the high-dimensional temporal structure of a human body, as validated in the literature (\eg, BeLFusion~\cite{DBLP:conf/iccv/BarqueroEP23}).
Built upon this insight, our method uses the encoder part of an autoencoder, largely borrowing its architecture from BeLFusion, for dimensionality reduction.

However, since a pretrained autoencoder is optimized for reconstructing original data, its latent space is not ideal for clustering.  
Therefore, UMAP is then used to further reduce the dimension for better clustering~\cite{DBLP:conf/icisp/AllaouiKC20,DBLP:journals/tvcg/EspadotoMKHT21}.

Our experiments verified that the latent space produced by our autoencoder\,+\,UMAP pipeline enables more stable clustering than the original motion space, the autoencoder latent space alone, and the UMAP embedding alone (see Supp.~D). 
The output dimensions of our autoencoder and UMAP are optimized based on the stability of clustering, which is defined with HDBSCAN used for the following clustering process; see Supp.~B for the details of this clustering stability.

\vspace*{-5mm}
\paragraph{Clustering in Motion Space.}

The low-dimensional latent space is then partitioned into multiple clusters, each representing a motion mode. This clustering is achieved by HDBSCAN~\cite{DBLP:journals/jossw/McInnesHA17,DBLP:conf/cvpr/KimWKGTK24,DBLP:conf/iccv/HassaniH19}, which is a density-based algorithm that identifies regions of higher sample density relative to their surroundings.
HDBSCAN requires no prior specification of the number of clusters and imposes no shape assumptions (in contrast, {\eg}, to k-means). It is therefore well-suited to our setting, where both the number of modes and the distribution of motions are unknown. 
Supp.~E.1 provides empirical evidence, confirming that HDBSCAN is the best among other clustering methods.

\subsection{Stage~2: Multimodality Metric Evaluation}
\label{subsection:MMCM_computation_for_predictions}

Recall that a multimodality metric must satisfy two requirements: \textbf{(a) coverage} and \textbf{(b) validity}.  
MMCM meets the requirement~(a) by relying on the explicit mode clustering. This requirement is quantified by the Mode Coverage Rate, denoted by \(C\).
MMCM also meets the requirement~(b) by checking whether each prediction belongs to any of the valid modes following the past motion, thereby assigning low scores to invalid motions. This is quantified by the Mode Validity Rate, denoted by \(V\).
Figure~\ref{fig:evaluation_metric_computation} illustrates the overall procedure that yields $C$, $V$, and MMCM.

To obtain $C$ and $V$ mentioned above, first of all, every predicted motion is assigned to one of the modes. This assignment is implemented as follows.
(P1) Each predicted motion $\hat{\mathbf{Y}}^{\,i}$ is concatenated with the last $N'_{p}$ frames of a given past motion.
To evaluate the smooth transition from the past to the future, these concatenated frames are fed into the autoencoder and UMAP to obtain the embedding feature in the low-dimensional space, as described in Sec.~\ref{subsection:clustering-based_mode_definition}.
(P2) Next, we compute the $\ell_2$ distance between this embedding and the centroid of each mode to assign the predicted motion to the nearest centroid’s mode.
However, even the nearest mode can be far from the predicted motion. Such a motion should be regarded as an abnormal motion ({\eg}, a motion that violates skeletal constraints or consists of discontinuously connected past and future sequences).
To remove the abnormal motions, we introduce a threshold~$\tau$: if the distance to the nearest mode exceeds~$\tau$, the prediction is labeled to be abnormal. $\tau$ is empirically set larger than the maximum distance observed when the same procedure is applied to only normal motions in the dataset (Supp.~I).
A set of modes to which all $I$ predicted motions are assigned is denoted as $\hat{\mathbf{M}} = \{\hat{m_1}, ..., \hat{m_I}\}$.
(P3) The aforementioned processes, P1 and P2, are also applied to all $K$ MMGTs.
A set of modes to which all the MMGTs are assigned is denoted as $\mathbf{M}=\{m_1,\dots,m_K\}$.
$\hat{\mathbf{M}}$ and $\mathbf{M}$ are used to obtain $C$ and $V$ as follows.

\vspace*{-3mm}
\paragraph{Mode Coverage Rate, $C$:}

The Mode Coverage Rate \(C\) is defined as follows:
\begin{equation}
C = \frac{\bigl\lvert \mathrm{uniq}(\mathbf{M}\cap \hat{\mathbf{M}}\bigr\})\bigr\rvert} {\bigl\lvert \mathrm{uniq}(\mathbf{M})\bigr\rvert},
\label{equation:C}
\end{equation}
where \(\operatorname{uniq}(\cdot)\) returns the set of unique elements. The denominator counts the modes of the MMGTs, while the numerator counts how many of those modes also include any predicted motions. Thus \(C\) quantifies the fraction of valid modes that the predicted motions cover. If all valid modes are covered, \(C=1\). On the other hand, if none is covered, \(C=0\). A higher \(C\) therefore indicates the broader coverage of the valid modes that could realistically follow the given past motion.

\vspace*{-3mm}
\paragraph{Mode Validity Rate, $V$:}

The Mode Validity Rate \(V\) is defined as follows:
\begin{equation}
\label{equation:V}
V = \frac{\bigl\lvert \{\,\hat{m_i} \mid \hat{m_{i}} \in {\mathbf{M}}\bigr\}\bigr\rvert} {I},
\end{equation}
where \(I\) is the number of predicted motions.
The numerator counts the predicted motions whose modes \(\hat{m}_{i}\) are included in the valid modes \(\mathbf{M}\).
\(V\) is the proportion of predicted motions that fall into valid modes. 
If all predicted motions are normal (\ie, included in the valid modes) and are outside of the valid modes, \(V = 1\) and \(V = 0\), respectively.

\vspace*{-3mm}
\paragraph{MMCM:}

The Mode Coverage Rate \(C\) and the Mode Validity Rate \(V\) are complementary.
\(C\) reflects the breadth of the covered modes, whereas \(V\) reflects the validity of individual predictions.
So, MMCM is defined as the harmonic mean of \(C\) and \(V\) as follows:
\begin{equation}
\label{equation:MultiModality}
\mathrm{MMCM}=\frac{2\,C\,V}{C+V}
\end{equation}
A higher MMCM indicates better multimodality.  
Viewing \(C\) as the recall term and \(V\) as the precision term, MMCM can be regarded as an F1-like score.

Algorithm~1 in Supp.~H summarizes the whole process mentioned in Sec.~\ref{section:method}.
This algorithm obtains the MMCM score of future motions predicted from a past motion.
The dataset-level MMCM is obtained by averaging over all test motions.

\section{Experiments}
\label{section:experiments}

\begingroup
\renewcommand{\arraystretch}{1.3}
\begin{table*}[t]
    \centering
        \caption{Quantitative comparison. \(C\) and \(V\) denote the average mode coverage rate and the average mode validity rate, respectively. \textcolor{red}{Red} and \textcolor{blue}{blue} indicate the best and second-best results for each category.}
        \label{table:result_baseline}
        \vspace*{-2mm}
        \setlength{\tabcolsep}{4pt}
        \begin{adjustbox}{width=\linewidth}
            \begin{tabular}{l|rrrrrr|rrrrrr}
            \hline
            \multirow{2}{*}{Method} & \multicolumn{6}{c|}{Human3.6M dataset} & \multicolumn{6}{c}{AMASS dataset} \\ \cline{2-13} 
                                    & \multicolumn{1}{c}{\textbf{MMCM↑ (C↑/V↑)}} & \multicolumn{1}{c}{APD↑} & \multicolumn{1}{c}{ADE ↓} & \multicolumn{1}{c}{FDE ↓} & \multicolumn{1}{c}{MMADE ↓} & \multicolumn{1}{c|}{MMFDE ↓}        & \multicolumn{1}{c}{\textbf{MMCM↑ (C↑/V↑)}} & \multicolumn{1}{c}{APD↑} & \multicolumn{1}{c}{ADE ↓} & \multicolumn{1}{c}{FDE ↓} & \multicolumn{1}{c}{MMADE ↓} & \multicolumn{1}{c}{MMFDE ↓} \\ \hline
            TPK & {\color{blue}{0.520}} (0.526/0.665) & 6.730 & 0.461 & 0.559 & 0.523 & \multicolumn{1}{r|}{0.568}  & 0.374 (0.661/0.352) & 9.277 & 0.656  & 0.676 & 0.675 & 0.674  \\
            DLow  & 0.487 (0.569/0.539) & 11.230 & 0.421 & 0.515 & 0.491 & \multicolumn{1}{r|}{0.528}  & 0.314 (0.657/0.265) & \color{blue}{13.169} & 0.590  & 0.611   & 0.618 & 0.616 \\
            GSPS  & 0.417 (0.520/0.441) & \color{blue}{14.755} & 0.393 & 0.500 & \color{blue}{0.478} & \multicolumn{1}{r|}{0.528}  & 0.256 (0.604/0.198) & 12.468 & 0.563  & 0.615  & 0.610 & 0.634 \\
            DivSamp & 0.409 (0.490/0.457) & \color{red}{15.558} & 0.381 & 0.500 & 0.486 & \multicolumn{1}{r|}{0.529}  & 0.198 (0.466/0.163) & \color{red}{24.724} & 0.565  & 0.647  & 0.624 & 0.667 \\
            HumanMAC  & 0.504 (0.471/0.737) & 6.235  & \color{blue}{0.369}   & 0.479   & 0.512     & \multicolumn{1}{r|}{0.546}  & -     & -      & -       & -     & -      & -         \\
            BeLFusion  & 0.509 (0.511/0.653)     & 7.618  & 0.372   & \color{blue}{0.472}   & \color{red}{0.473}     & \multicolumn{1}{r|}{\color{red}{0.507}} & {\color{red}{0.386}} (0.637/0.382)  & 9.378  & \color{blue}{0.513}   & \color{blue}{0.561}   & \color{red}{0.569}  & \color{blue}{0.586}     \\
            CoMusion  & {\color{red}{0.521}} (0.528/0.666) & 7.644 & \color{red}{0.351} & \color{red}{0.459}  & 0.495     & \multicolumn{1}{r|}{\color{blue}{0.509}} & {\color{blue}{0.378}} (0.642/0.403)     & 10.842 & \color{red}{0.494} & \color{red}{0.547} & \color{blue}{0.579} & \color{red}{0.585}  \\ \hline
            \end{tabular}
        \end{adjustbox}
        \vspace*{-2mm}
\end{table*}
\endgroup

\subsection{Experimental Setup}
\label{subsection:experimental_setup}

\noindent
\textbf{Datasets.} 
All experiments are conducted on Human3.6M dataset (H36M)~\cite{DBLP:journals/pami/IonescuPOS14} and AMASS dataset~\cite{DBLP:conf/iccv/MahmoodGTPB19}. H36M is a motion capture dataset comprising 3.6M frames (50 Hz) from 11 subjects. Following common practice~\cite{DBLP:conf/eccv/YuanK20,DBLP:conf/iccv/BarqueroEP23,DBLP:conf/eccv/SunC24,DBLP:conf/iccv/ChenZLPXL23}, we use the 17-keypoints and set the past window to \(B=25\) frames and the prediction window to \(T=100\) frames.
AMASS is a large-scale dataset that merges 22 motion-capture datasets
~\cite{AMASS_ACCAD, AMASS_BMLhandball, AMASS_BMLmovi, AMASS_BMLrub, AMASS_CMU, AMASS_DanceDB, AMASS_DFaust, AMASS_EyesJapanDataset, AMASS_GRAB, AMASS_HDM05, AMASS_HUMAN4D, DBLP:journals/ijcv/SigalBB10, AMASS_KIT-CNRS-EKUT-WEIZMANN, AMASS_MoSh, AMASS_PosePrior, AMASS_SFU, AMASS_SOMA, AMASS_TCDHands, AMASS_TotalCapture} with 9M frames (60 Hz). We use the 21-keypoints, \(B=30\) frames of the past window, and \(T=120\) frames of the prediction window following~\cite{DBLP:conf/iccv/BarqueroEP23, DBLP:conf/eccv/SunC24}.

\noindent
\textbf{Hyper parameters.} 
For the MMCM hyper parameters, the autoencoder compresses 4,944-dimensional (H36M) and 7,749-dimensional (AMASS) motions to 64 dimensionals, and UMAP~\cite{DBLP:journals/jossw/McInnesHSG18} further reduces it to 2 dimensions. For HDBSCAN~\cite{DBLP:journals/jossw/McInnesHA17}, we tune min\_cluster\_size, which is the minimum cluster size, and min\_samples, which controls noise sensitivity. These parameters are set to (15, 1) for H36M and (50, 1) for AMASS. All values are chosen to balance cluster persistence and noise rate; experimental results to explore these parameters are shown in Supp.~B. 
For the MMGT parameters, we follow previous studies~\cite{DBLP:conf/eccv/YuanK20,DBLP:conf/iccv/BarqueroEP23,DBLP:conf/eccv/SunC24}, we using a past-window size of 1 frame and a similarity threshold of 0.5 (H36M) / 0.4 (AMASS).

\noindent
\textbf{HMP Metrics.} 
Our proposed MMCM is compared with APD, which is the prevailing metric for diversity.  
For reference, we also include the accuracy scores, \ie, ADE, FDE, MMADE, and MMFDE.

\noindent
\textbf{HMP methods.}
HMP metrics are evaluated with the following existing HMP methods:  
CoMusion~\cite{DBLP:conf/eccv/SunC24},  
BeLFusion~\cite{DBLP:conf/iccv/BarqueroEP23},  
HumanMAC~\cite{DBLP:conf/iccv/ChenZLPXL23},
DivSamp~\cite{DBLP:conf/mm/DangNLZL22},
GSPS~\cite{DBLP:conf/iccv/MaoLS21},
DLow~\cite{DBLP:conf/eccv/YuanK20} and
TPK~\cite{DBLP:conf/iccv/WalkerMGH17}.
The HMP metric scores of all these methods are obtained with the codes and weights distributed by their authors.
One exception is that, since HumanMAC weights for AMASS are not publicly available, we report HumanMAC results only on H36M.
All methods predict $I=50$ future motions for each past sequence.

\subsection{Quantitative Comparison}
\label{subsection:baseline_comparison}

Table~\ref{table:result_baseline} shows the quantitative results of six metrics, \ie, MMCM, APD, ADE, FDE, MMADE, and MMFED.
Section~\ref{subsection:baseline_comparison} focuses on MMCM and APD, each of which evaluates whether a variety of motions are predicted.
In both H36M and AMASS, MMCM and APD exhibit markedly different evaluation trends.
To analyze the reason why MMCM and APD yield the different trends, Sec.~\ref{subsection:baseline_comparison} focuses on the H36M results.

\begin{figure}[t]
  \begin{center}
     \includegraphics[width=\linewidth]{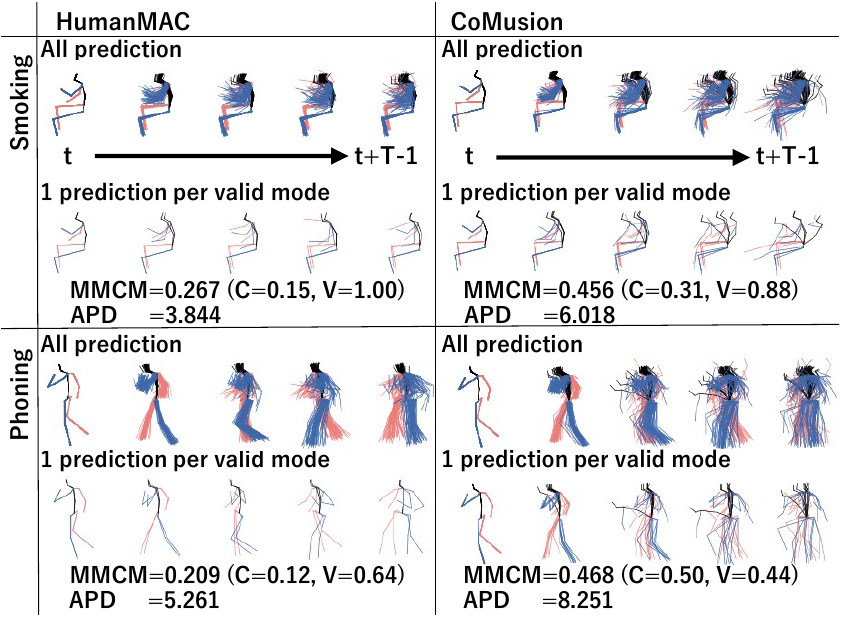}
     \vspace*{-7mm}
     \caption[Multimodality comparison between HumanMAC and CoMusion.]{Multimodality comparison between HumanMAC and CoMusion. The left and right columns show motions predicted by HumanMAC and CoMusion, respectively. In each of the smoking and phoning examples, the top row overlays all 50 predictions, while the bottom row extracts one representative motion per valid mode detected by MMCM.
     }
     \label{fig:humanmac_vs_comusion}
   \end{center}
\end{figure}

\textbf{Can MMCM measure (a) coverage?}  
Comparing HumanMAC with CoMusion, it is observed that CoMusion attains a higher MMCM, which stems from its larger Mode Coverage Rate. To understand why, Fig.~\ref{fig:humanmac_vs_comusion} compares the predictions of HumanMAC and CoMusion for the same past motion.
Two examples (\ie, ``smoking'' and ``phoning'' motions) are shown.
For each example, the top row visualizes all $I=50$ predicted future motions, while the bottom row shows one sample motion randomly extracted from each valid mode for better visibility.
In both examples, all 50 future motions predicted by HumanMAC occupy only two valid modes (\ie, \(C = 0.15\) and $C=0.12$ in these smoking and phoning examples, respectively). In addition, these two modes include only similar body poses (\ie, ``upright sitting poses'' and ``upright walking poses'' in these smoking and phoning examples, respectively). CoMusion, by contrast, outputs various modes (\ie, \(C = 0.31\) and $C=0.50$ in these smoking and phoning examples, respectively).
MMCM scores shown below each sequence qualitatively validates that CoMusion provides richer multimodality.
These observations validate that MMCM can successfully measure (a) coverage.
The Supp. provides additional results under the same experimental settings in D.1. 

\begin{figure}
  \begin{center}
     \includegraphics[width=\linewidth]{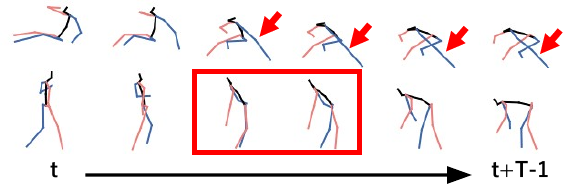}
     \vspace*{-6mm}
     \caption[Outputs from DLow classified as abnormal.]{Outputs from DLow classified as abnormal by MMCM. Top: an abnormal motion in which the arms become unrealistically elongated. Bottom: an abnormal motion where the torso is horizontally mirrored from right to left.}
     \label{fig:outliner_dlow}
   \end{center}
   \vspace*{-6mm}
\end{figure}

\textbf{Can MMCM measure (b) validity?}
In Table~\ref{table:result_baseline}, DLow records a higher APD score (\ie, 11.230).
However, the MMCM score of DLow is low (\ie, 0.487) because its Mode Validity Rate is low (\ie, \(V = 0.539\), which is the third from the bottom), while its Mode Coverage Rate is high (\ie, \(C = 0.569\), which is the top). The harmonic mean of $V$ and $C$ gets lower.
Abnormal motions excluded from valid modes, which decrease $V$ as mentioned above, are shown in Fig.~\ref{fig:outliner_dlow}. In the upper example, the arms stretch to invalid lengths, while in the lower example, the torso flips unnaturally from facing right to facing left, violating skeletal constraints.
In the upper sample, the length from left shoulder to left hand changes from 0.55\,m to 0.94\,m, revealing a quantitative abnormality. For this sequence, we computed MMCM and APD after removing such length-abnormal motions (\ie, motions where any inter-joint length changes by at least 50\% relative to the past). The results show that MMCM correctly increases (0.597$\rightarrow$0.624), indicating that the abnormality-free predictions are judged better. In contrast, APD erroneously decreases (11.016$\rightarrow$10.011), implying a preference for the set containing abnormal motions. 
These results show that MMCM fulfills requirement (b) validity, while APD provides the opposite result.
More qualitative results under the same experimental settings are shown in Supp.~E.2.
More quantitative evidences are provided in Sec.~\ref {subsection:noised_motion_result}.

\subsection{Robustness to Abnormal Motion Sequences}
\label{subsection:noised_motion_result}

\begin{figure}
  \begin{center}
     \includegraphics[width=\linewidth]{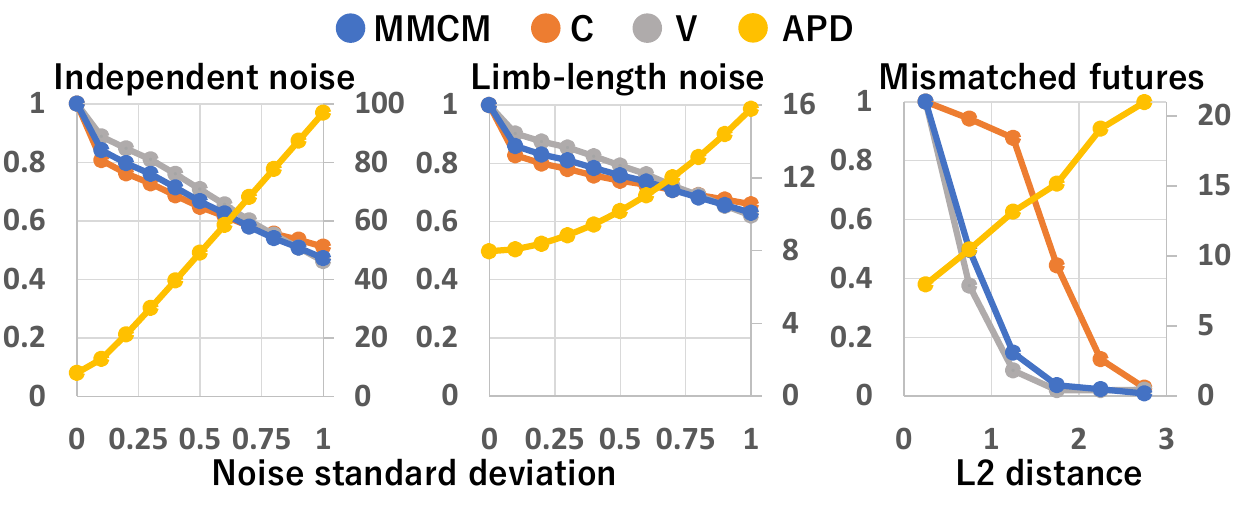}
     \vspace*{-6mm}
     \caption[Sensitivity of the metrics to synthetic abnormal motions.]{Comparison of MMCM (left axis) and APD (right axis) with three kinds of abnormal data. Left: independent Gaussian noise added to every joint position. Center: bone-length noise produced by randomly scaling bones. Right: mismatched past and future motions. In both MMCM and APD, higher is better. As anomaly level rises (\ie, rightward on the horizontal axis), APD erroneously increases (rewarding abnormal motions, which is undesirable for a good metric), whereas MMCM correctly decreases.
     %(rewarding normal motions).
     }
     \label{fig:result_noise_motion}
   \end{center}
   \vspace*{-3mm}
\end{figure}

\begin{figure*}[!t]
  \begin{center}
     \includegraphics[width=\linewidth]{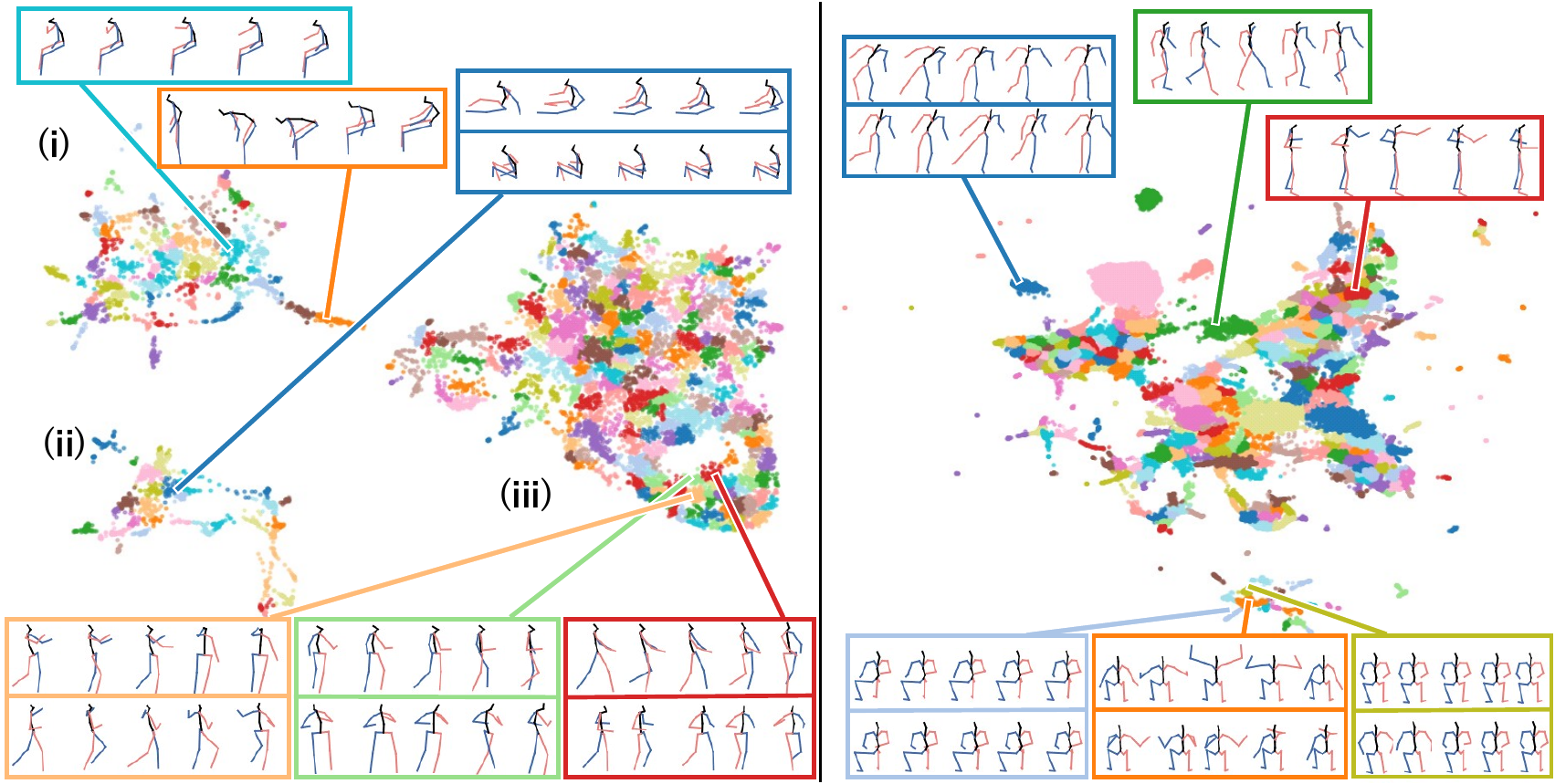}\\
     (a) Human3.6M \hspace*{70mm} (b) AMASS
     \caption[Clustering-based mode definition.]{Clustering-based mode definition. An embedding of the latent motion space is partitioned into color-coded modes. Similar motions cluster together while different motions are well separated, confirming the validity of the mode definition.}
     \label{fig:mode_definition}
   \end{center}
\end{figure*}

To investigate more rigorously whether (b) validity is satisfied and unsatisfied by MMCM and APD, respectively, we conducted two additional experiments on the modified H36M where future motions are abnormal.
With such abnormal future motions, a good metric should get worse.
The following experiments verify whether or not the diversity metric (\ie, APD) and multimodality metric (\ie, MMCM) can correctly get worse with the abnormal future motions.

\noindent
\textbf{Noise injection.}
In this experiment, we evaluate the kinematically abnormal motions.
Starting from MMGT future motions, noisy motions that are kinematically abnormal are generated with each of the following two perturbation types:
(i) Independent noise, where zero-mean Gaussian noise is added to every joint independently, and (ii) limb-length noise, where the length of selected bones is randomly scaled, producing invalidly long or short limbs.  
Experimental results with these two types of noise are shown in Fig.~\ref{fig:result_noise_motion}.
For both noise types, APD rises almost linearly with the perturbation magnitude.
In contrast, MMCM correctly decreases as the noise standard deviation increases, showing that it penalizes noise-corrupted abnormal motions.  

\noindent
\textbf{Mismatched futures.}  
In this experiment, we evaluate motions, in each of which their past and future sequences are kinematically valid independently but connected discontinuously.
Such a motion is synthesized by connecting each past sequence (denoted by $M_{p}$) to any future sequence (denoted by $M_{f}$) in the dataset.
All of these synthesized sequences are grouped depending on the discontinuity.
This discontinuity is represented by the $\ell_2$ distance between the last frames of $M_{p}$ and the real past sequence followed by $M_{f}$.
In the graph shown in Fig.~\ref{fig:result_noise_motion} (iii), this discontinuity is along the horizontal axis. 
It can be seen that APD grows with the anomaly level, whereas MMCM correctly decreases so that the requirement~(b) is satisfied.

\subsection{Clustering Results (Mode definition Results)}
\label{subsection:clustering_result}

This section evaluates the validity of the motion mode definition via clustering.  
The procedure proposed in Sec.~\ref{subsection:clustering-based_mode_definition} partitions the motion space into 324 modes on H36M and 278 modes on AMASS.
Figure~\ref{fig:mode_definition} visualizes the clustered latent spaces together with representative motions.

A sound mode definition should neither over-segment nor under-segment. That is, similar motions should belong to the same mode, while clearly different motions should belong to different modes. 

\noindent
\textbf{Human3.6M.}
The scatter plot reveals three regions: 
(i) the upper left, containing chair-sitting motions,  
(ii) the lower left, containing floor-sitting or lying motions, and  
(iii) the right side, containing standing and walking motions. 
These three regions were found to be divided by very different types.
For more detailed verification, adjacent modes are inspected.
Two motions are sampled from each of the three adjacent modes and shown at the bottom of the figure. 
Those sampled from the same mode are enclosed by the rectangles of the same color (\ie, \textcolor{red}{red}, \textcolor{green}{green}, or \textcolor{orange}{orange}).
While all of these three modes include turning motions, their turning characteristics differ: (1) \textcolor{red}{the red mode} turns from right-facing to front-facing, (2) \textcolor{green}{the green mode} turns from front toward the right, and (3) \textcolor{orange}{the orange mode} turns from right toward depth.
These samples validate that our clustering process achieves a fine partitioning of motions.

\noindent
\textbf{AMASS.}
The mode definition of AMASS in Fig.~\ref{fig:mode_definition} (b) also shows that similar motions are assigned to the same mode, whereas motions from different modes have different characteristics. This motion distribution can be confirmed in three adjacent modes shown at the bottom, which are colored \textcolor{cyan}{cyan}, \textcolor{orange}{orange}, and \textcolor{olive}{olive}.

\subsection{Comparison of MMCM vs. MMADE/MMFDE}
\label{section:cpmarison_with_mmade}

MMADE and MMFDE, used in the literature~\cite{DBLP:conf/eccv/YuanK20}, can be regarded as accuracy and diversity metrics because they use MMGTs.
However, MMCM is superior to MMADE and MMFDE in terms of measuring multimodality
due to the following reasons:
(a) coverage: MMADE and MMFDE focus on major modes that contain many MMGTs, thereby underestimating rare modes, and
(b) validity: MMADE and MMFDE ignore inaccurate predictions.

\begin{figure}
  \begin{center}
     \includegraphics[width=\linewidth]{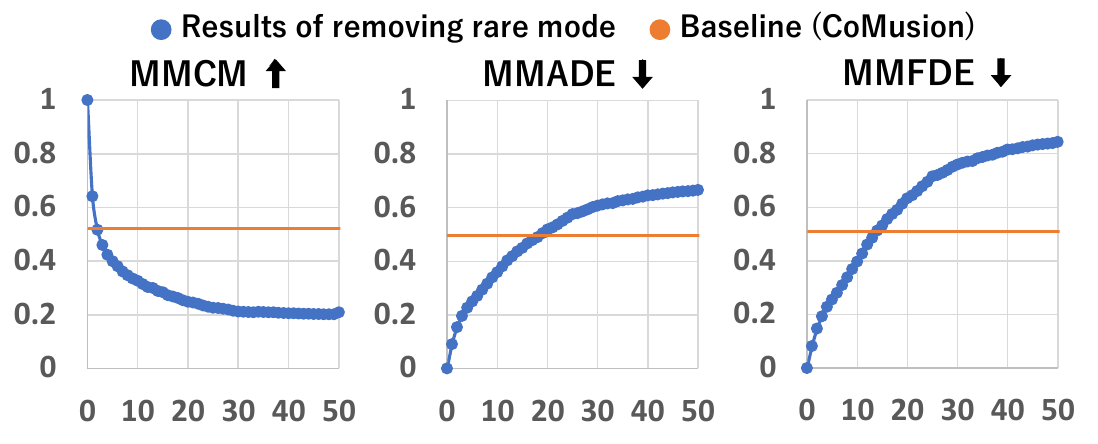}
     \vspace*{-6mm}
     \caption[Comparison of MMCM and MMADE by removing rare mode.]{Comparison of MMCM, MMADE, and MMFDE by removing rare modes. The horizontal axis indicates the percentage of being rare. Compared with MMADE and MMFDE, MMCM penalizes predictions that omit rare modes much more sharply.
     }
     \label{fig:suppl_mmade_raremode}
   \end{center}
\end{figure}

\noindent
{\bf (a) Coverage.}
Starting from all MMGT future motions used as predictions, we progressively remove the predictions that belong to rare modes having fewer predictions.
The results of how MMCM and MMADE/MMFDE change as these rare modes are progressively removed are shown in Fig.~\ref{fig:suppl_mmade_raremode}.
The horizontal axis of each graph in Fig.~\ref{fig:suppl_mmade_raremode} indicates the percentage of predictions included in each mode. If the horizontal axis is at $v$, predictions included in modes with $v$ \% or less predictions are removed to evaluate the metric score, shown on the vertical axis.
If the percentage is~0 (no removal), all metrics achieve their best scores.
As the percentage grows, all the scores get worse, suggesting that all metrics successfully penalize the loss of even rare modes.
However, compared with MMCM, the scores of MMADE and MMFDE get worse slowly.
Thus, MMADE and MMFDE are not sensitive to a change in (a) coverage, whereas MMCM responds appropriately to the loss of even rare modes.
Note that predicting not only major modes but also rare modes is crucial for safety-critical applications.

\begin{figure}
  \begin{center}
     \includegraphics[width=\linewidth]{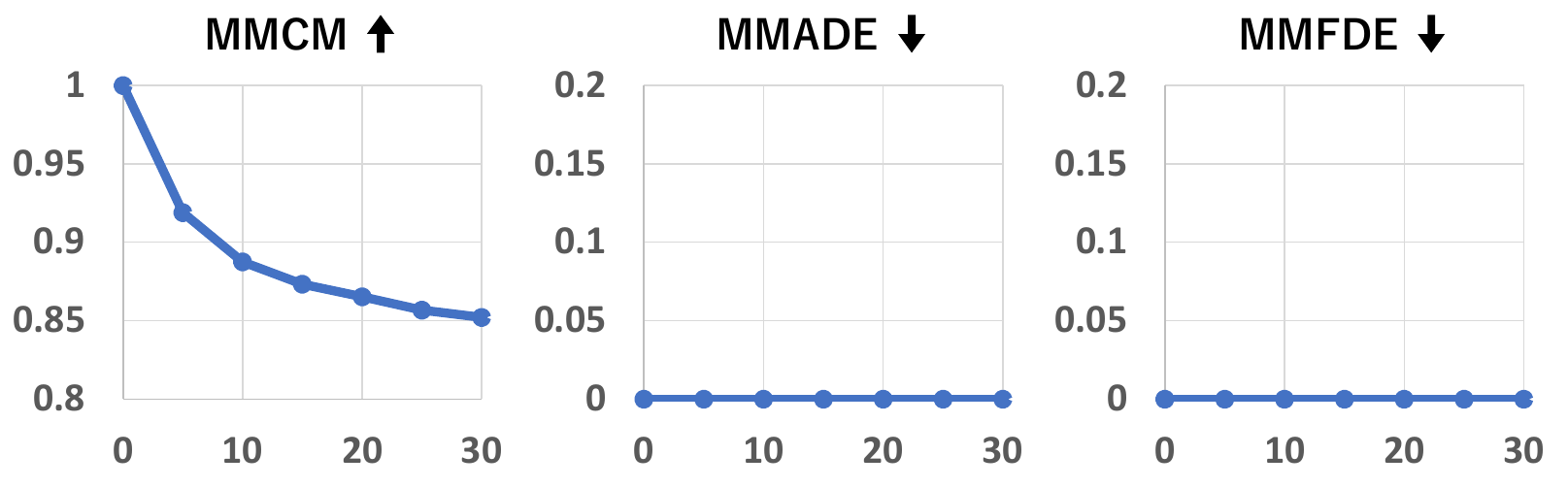}
     \vspace*{-6mm}
     \caption[Comparison of MMCM and MMADE by adding noisy motion.]{Comparison of MMCM, MMADE, and MMFDE by adding noise to MMGT futures. The horizontal axis shows the number of added noisy motions. As the number of noisy motions increases, MMCM decreases, whereas MMADE/MMFDE stay at their best values, ignoring the noisy motions.}
     \label{fig:mmade_noise}
   \end{center}
\end{figure}

\noindent
{\bf (b) Validity.}
Starting from all MMGT future motions used as predictions, we progressively add noisy motions to these predictions to observe how MMCM, MMADE, and MMFDE respond.  
The results are shown in Fig.~\ref{fig:mmade_noise}, in which the horizontal axis indicates the number of noisy motions.  
With no noisy motions, all metrics achieve their best scores.  
As the number of noisy motions increases, however, MMADE and MMFDE remain unchanged at their best scores, treating the prediction set with the noisy motions as perfectly accurate. On the other hand, MMCM degrades steadily. 
These results demonstrate that our MMCM can capture (b) validity, while MMADE and MMFDE cannot.

\subsection{Computational Efficiency}
\label{section:computationnal_efficiency}

For MMCM, the time per prediction on a CPU was 0.06\,s on H36M and 0.07\,s on AMASS. These times are sufficiently fast for practical use of offline evaluation.

\subsection{Robustness with Respect to Hyperparameters}
\label{section:robustness_hyperparameter}

\begin{figure}
  \begin{center}
     \includegraphics[width=\linewidth]{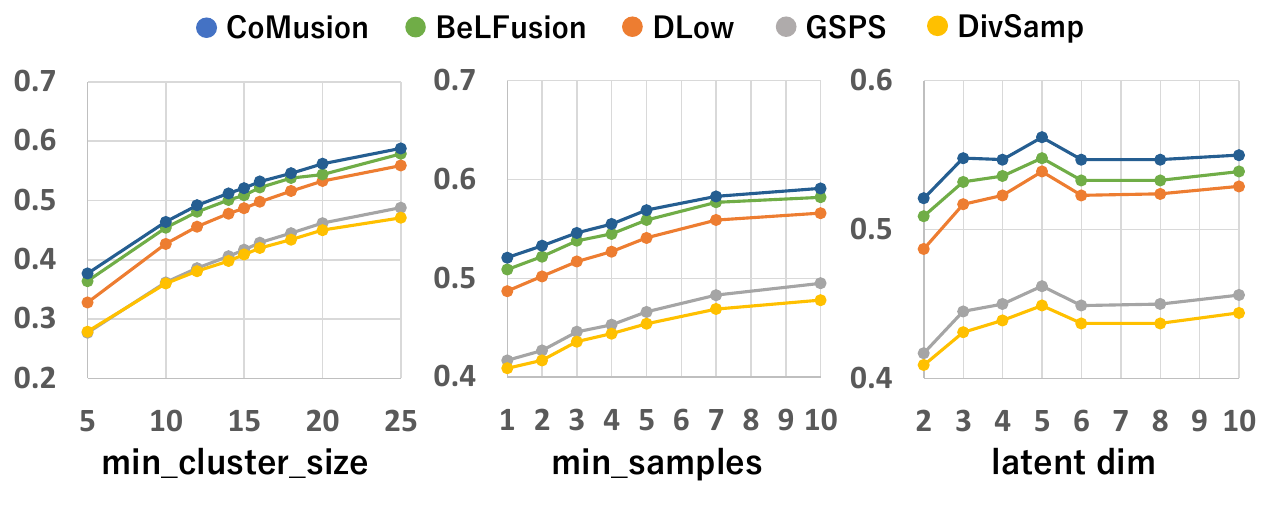}
     \vspace*{-6mm}
     \caption[MMCM sensitivity to the some hyper parameters.]{MMCM sensitivity to the some hyper parameters on H36M. For all hyperparameters, changing our selected settings does not affect the relative ranking of HMP methods.}
     \label{fig:robustness_hyperparameter_h36m}
     \vspace*{-4mm}
   \end{center}
\end{figure}

We examine how MMCM’s hyperparameters affect evaluation results and demonstrate the metric’s robustness to hyperparameter choices on H36M. The hyperparameters of MMCM include the latent dimensionality and the HDBSCAN clustering parameters ``min\_cluster\_size'' and ``min\_samples''. We conducted comprehensive experiments on all of them. Specifically, the latent dimensionality was controlled by changing the output dimension of UMAP. Fig.~\ref{fig:robustness_hyperparameter_h36m} shows how the MMCM scores of the HMP methods (CoMusion, BeLFusion, DLow, GSPS, DivSamp) change as these hyperparameters are varied.
Across all hyperparameters, changing their values slightly affects the absolute MMCM scores but not the relative ranking of methods. Since a key requirement for an evaluation metric is to discriminate between methods, these results indicate that MMCM is robust to hyperparameter settings. Results on AMASS under the same conditions appear in Supp.~C.

\section{Conclusion}
\label{section:conclusion}

This paper proposed MMCM, a new metric that measures multimodality in HMP.
Unlike the prevailing diversity metric like APD, MMCM captures multimodality in two complementary ways: (a) using a clustering–based mode definition, MMCM explicitly evaluates whether the predictions cover multiple modes; (b) MMCM evaluates whether each motion belongs to a valid mode given a past.
Our experiments revealed that even state-of-the-art HMP methods attain low MMCM scores, as shown in Table~\ref{table:result_baseline}.
We anticipate that MMCM will facilitate future HMP research directions.

\textbf{Limitations and future work.} MMCM relies on an autoencoder, UMAP, and HDBSCAN. Therefore, evaluating a new dataset requires new training and hyper parameter tuning, while automatic selection of the hyper parameters is also proposed in our work; see Supp.~B.
By converting arbitrary keypoint layouts to H36M/AMASS format (\eg,~\cite{DBLP:conf/iccv/MahmoodGTPB19}), we can directly leverage our pre-defined modules.
Another direction for future work is to evaluate HMP results obtained from estimated human poses~\cite{DBLP:conf/cvpr/TripathiMHTBT23,DBLP:conf/cvpr/Dwivedi0PFB24,DBLP:conf/cvpr/VendrowL0R23,DBLP:conf/wacv/TaketsuguU24} , instead of motion capture data~\cite{DBLP:journals/pami/IonescuPOS14,DBLP:conf/iccv/MahmoodGTPB19}, in more realistic scenarios.

{
    \small
    \bibliographystyle{ieeenat_fullname}
    \bibliography{main}
}

\end{document}